# Image Captioning based on Deep Reinforcement Learning

Haichao Shi
Institute of Information Engineering,
Chinese Academy of Sciences
School of Cyber Security, University of
Chinese Academy of Sciences
Beijing, China
shihaichao@iie.ac.cn

Peng Li[†]
College of Information and Control
Engineering, China University of
Petroleum (East China)
Qingdao, China
lipeng@upc.edu.cn

Bo Wang[†]
National Computer Network
Emergency Response Technical
Team/Coordination Center of China
Beijing, China
wangbo@cert.org.cn

Zhenyu Wang
Institute of Information Engineering,
Chinese Academy of Sciences
Beijing, China
wangzhenyu@iie.ac.cn

## ABSTRACT

Recently it has shown that the policy-gradient methods for reinforcement learning have been utilized to train deep end-to-end systems on natural language processing tasks. What's more, with the complexity of understanding image content and diverse ways of describing image content in natural language, image captioning has been a challenging problem to deal with. To the best of our knowledge, most state-of-the-art methods follow a pattern of sequential model, such as recurrent neural networks (RNN). However, in this paper, we propose a novel architecture for image captioning with deep reinforcement learning to optimize image captioning tasks. We utilize two networks called "policy network" and "value network" to collaboratively generate the captions of images. The experiments are conducted on Microsoft COCO dataset, and the experimental results have verified the effectiveness of the proposed method.

**ACM Reference format:**

Haichao Shi, Peng Li, Bo Wang and Zhenyu Wang. 2018. Image Captioning based on Deep Reinforcement Learning. SIG Proceedings Paper in word Format. In *Proceedings of the 10th International Conference on Internet Multimedia Computing and Service, Nanjing, China, August 2018 (ICIMCS 2018)*, 5 pages.

## 1 INTRODUCTION

Learning based methods have been widely used in various image analysis tasks [30,31,32,33,34,35,36]. Image captioning is a challenging task of generating the natural language description for the input image. It requires a fine-grained understanding of the global and the local entities in an image, as well as the relationships and attributes. Image captioning tasks have attracted increasingly interests in computer vision recently. Most state-of-the-art methods [2,3,4,5,6,7] choose to utilize the encoder-decoder framework to generate the captions of images. Inspired by the deep learning approaches, which have yielded impressive results on the computer vision tasks, the encoder-decoder models usually utilize the convolutional neural networks to encode the images and employ the recurrent neural networks to decode the semantic information to integrate complete sentences. These models are trained end-to-end using back-propagation neural networks, and have achieved fairly good results on Microsoft COCO dataset [1].

In this paper, we introduce a novel architecture with deep reinforcement learning for image captioning. Different from the former works, which are learned to train a recurrent model to look for the next suitable word, we utilize two networks called "policy network" and "value network" to jointly learn to predict the correct words at each state. In detail, the policy network is utilized to evaluate the confidence of predicting the next word according to the current state. The value network is utilized to evaluate the reward value of the predictions of the current state. In other words, the value network aims at adjusting the targets of predicting the linguistic of the images towards generating captions as natural as possible. Based on the deep reinforcement learning, our model can generate captions similar to human natural language descriptions using the two networks. As is shown in Fig. 1, is an instance of the proposed image caption model based on deep reinforcement learning. Based on reinforcement learning, the policy network gives some possible actions about the object. Then, the value network makes decisions on whether to choose the action given by the policy network based on the evaluated reward scores. Both networks are devoted to satisfying the ultimate goal of generating a good enough description.

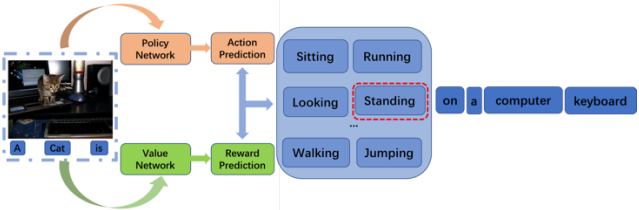

**Figure 1: An instance of the proposed image caption model based on deep reinforcement learning. The policy network is intended to predict the action of the object according to the current state, and the value network is willing to make an inference based on the rewards.**

We conduct detailed experiments to demonstrate the comparative performance than the state-of-the-art approaches. Our evaluation indicators include BLEU [22], METEOR [23], CIDEr [24] and ROUGE [25]. The contributions of this paper are summarized as follows:

- We integrate the deep reinforcement learning with image caption tasks to generate descriptions more like natural language. We use two networks called "policy network" and "value network" to complement with each other to improve the quality of generated descriptions.
- We optimize the two networks using temporal-difference (TD) method [26]. As for the training process, we use supervised learning with cross entropy loss to train the sequence part of policy network and with mean squared loss to train the value network.

The rest of the paper is structured as follows. In Section 2, we discuss the related work of image caption and reinforcement learning. In Section 3, we elaborate the proposed method. In Section 4, experiments are conducted to demonstrate the effectiveness of the proposed method. In Section 5, we draw conclusions.

## 2 RELATED WORK

### 2.1 Image Captioning

Image caption is a comprehensive problem that integrates computer vision, natural language processing and machine learning. With the rise of machine translation and big data, most state-of-the-art methods follow an encoder-decoder framework to generate captions for natural images. To the best of our knowledge, the encoder is always a convolutional neural network, using the characteristics of the last fully connected layer or convolutional layer as the features of an image. The decoder is generally a recurrent neural network and is mainly utilized for generating image description. Due to the problem of gradient descent in ordinary RNNs, RNN can only memorize the contents of the previous limited time unit. Then comes the LSTM (Long Short Term Memory) [27], which is a special RNN architecture that can solve problems such as the vanish of gradients and has long-term memory. Therefore, the LSTM is gradually used in the decoder stage.

In Vinyals's [5] work, they proposed an encoder-decoder framework, which utilizes convolutional neural networks to extract image features and generates target language description through LSTM to maximize the maximum likelihood estimation of target description. While in Fang's [8] work, they utilized multiple instance learning to train the detector to extract words contained in an image at first, then a statistical model is learned to generate descriptions. In Xu's [9] work, they incorporated the attention mechanism with image caption. They propose to combine spatial attention mechanisms in the convolutional features of the image, and input context information into the encoder-decoder framework. These works all combine CNN (convolutional neural network) and RNN (recurrent neural network) to do caption generation tasks. The words are sequentially drawn according to the local confidence. Such methods usually choose the words with top local confidence. As a result, some good caption results may be missed. In contrast, our model can choose the suitable caption results via the reward scores to generate a good description.

### 2.2 Reinforcement Learning

Reinforcement learning is usually the core problem in computer gaming, control theory and path planning, etc. These problems all meet with the same conditions, there exists agents that need to interact with the environment, execute a series of actions and are intended to complete the expected goals. Before the rise of the deep reinforcement learning, some preliminary work has been carried out. However, these tasks use only deep neural networks to reduce the dimension of high-dimensional input data so that the traditional reinforcement learning algorithm can process it. Riedmiller et al. [14] proposed to use a multilayer perceptron to approximate represent Q-value function, and a Neural Fitted Q Iteration (NFQ) algorithm was proposed. In Lange's work [15], they combined deep learning model with reinforcement learning to propose a Deep Auto-Encoder (DAE). Abtahi et al. [16] utilized deep belief network to be used as an approximator in the traditional reinforcement learning, which greatly improves the learning efficiency of the agent and is successfully applied to character segmentation tasks of license plate images. Recently,

Image Captioning based on Deep Reinforcement Learning

Silver et al. [17] designed a professional-level computer Go program using deep neural networks and Monte Carlo Tree Search. Human-level gaming control [18] was achieved through deep Q-learning. A visual navigation system [21] was proposed recently based on actor-critic reinforcement learning model. There are also having generation tasks using reinforcement learning, such as [19]. They use reinforcement learning [20] to train its model to generate texts by directly optimizing a user-specified evaluation metric. In this paper, we do generation tasks on image captioning using deep reinforcement learning.

## 3 Image Captioning based on Deep Reinforcement Learning

In this section, we first elaborate the formulation of image captioning based on deep reinforcement learning. Then we introduce our model architecture and the training procedure.

### 3.1 Problem Formulation

Since we introduce deep reinforcement learning to image caption tasks, we formulate the problem into the scheme of reinforcement learning. Thus, we model this problem as a decision-making process. There are four factors affecting the whole process, agent, action, environment and goal. All the four factors have an impact with each other. The agent is intended to interact with the environment, and executes a series of actions to achieve the goal. The evaluation indicators are formulated according to the rewarding mechanism. In the task of image captioning, given an image $I$, to generate a natural description $S = \{\omega_1, \omega_2, \dots, \omega_n\}$ about it, where $\omega_i$ is a word in the description and $n$ is the number of words. What's more, our model includes a policy network $q_\pi$ and a value network $v_\pi$. In this situation, the two networks can be viewed as agents, the image $I$ and the generated description $S_t = \{\omega_1, \omega_2, \dots, \omega_t\}$ can be regarded as environment, which $t$ is the internal time. The prediction of the next word $\omega_{t+1}$ is viewed as an action.

### 3.2 Model Architecture

*3.2.1 The Policy Network.* Similar with the architecture of encoder-decoder, the policy network $q_\pi$ consists of two networks, a Convolutional Neural Network (CNN) and a Recurrent Neural Network (RNN), which provides a probability for the agent to take actions at each state. The architecture of policy network $q_\pi$ is shown in Fig. 2. We make a hypothesis that, the current state is $s_t$, which include the environment $e = \{I, \omega_1, \omega_2, \dots, \omega_t\}$ to interact with. The action is $a_t = \omega_{t+1}$. When the policy network is fed into an image, the CNN is utilized to encode the visual information. Then the information is fed into a RNN module, providing the action $a_t$ at each step according to the hidden state $h_t$. For the RNN is able to keep the sequential information, the $\omega_t$ generated by it at time $t$ will be fed back into RNN at next step. When the inputs update, the hidden state will also be updated to the next state. Empirically, we design the function of $q_\pi$ for each input $x_t$ by the following equations:

$$h_t = RNN(h_{t-1}, x_t) \quad (1)$$
$$x_0 = W_x CNN(I) \quad (2)$$
$$x_t = \phi(\omega_{t-1}), t \geq 1 \quad (3)$$
$$q_\pi(a_t|s_t) = \varphi(h_t) \quad (4)$$

Where $W_x$ is the weight of the input of convolutional neural network. $x_0$ is the initial input of RNN. $\phi$ and $\varphi$ represent the input and output of the RNN.

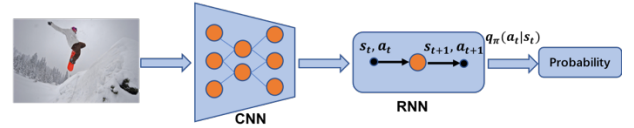

**Figure 2: The architecture of policy network $q_\pi$, which is comprised of a CNN and a RNN. Through the policy function $q_\pi(a_t|s_t)$, the probability of executing an action $a_t$ at a certain state $s_t$ is computed.**

*3.2.2 The Value Network.* For the value network, it contains three parts, a CNN, a RNN and a Linear Mapping Layer, here we use a perceptron model, which is utilized to evaluate the predictions to choose the most suitable action. As is shown in Fig. 3, is the architecture of value network $v_\theta$. The CNN is utilized to encode the visual information of the given image, the RNN is utilized to encode the semantic information of the partially given caption. And the Linear Mapping Layer is designed to predict the generated captions to give a reward on the caption.

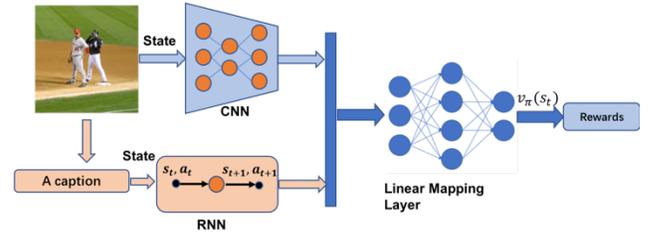

**Figure 3: The architecture of value network $v_\pi$, which is comprised of a CNN, a RNN and a Linear Mapping Layer. The value network can evaluate the policy's value given an image and a partially generated caption.**

### 3.3 Reward Mechanism and Training Strategy

*3.3.1 Reward Mechanism.* The reward mechanism is the measurement of how well the action is performing to complete the goal. We learn from the Reproducing Kernel Hilbert Space (RKHS) [28] to map the raw data through a non-linear mapping method, which makes the original linearly inseparable problem become a linearly separable problem. In our model, we utilize a linear mapping method to map the images and captions into a semantic embedding space, where we can calculate the distance between the images and captions. For the value network $v_\pi$ aims at giving a certain reward to an action, we denote the components of $v_\pi$ as $CNN, RNN, l_m$. Given an image caption $S =$ $\{\omega_1, \omega_2, ..., \omega_T\}$, its embedding feature is represented as $h_{T-1}(S)$, which depends on the last state of $h_T$. $f_I$ represents the feature extracted by the CNN and $l_m$ represents the mapping function of mapping the features into the embedding space. For the mapping loss, we can define it as follows:

$$L_m = \sum_{f_I} \sum_S \gamma [\max(0, h_{T-1}(S) \cdot l_m(f_I)) - h_T(S) \cdot l_m(f_I)] \quad (5)$$

Where $\gamma$ is the penalty coefficient varying from (0,1), $h_T$ and $h_{T-1}$ are the adjacent hidden states. And we define the final reward as follows:

$$R_T = \frac{h_{T-1}(S) \cdot l_m(f_I)}{||h_{T-1}(S) \cdot l_m(f_I)||} \quad (6)$$

Then the final loss can be represented as follows:

$$L_R = \alpha(L_m + R_T) \quad (7)$$

Where $\alpha$ is varying from (0,1).

*3.3.2 Training Strategy.* Firstly, we train the two networks respectively. The policy network attempts to get the possible action predictions. We train the policy network utilizing supervised learning and optimize it with cross entropy loss. Then we train the value network by minimizing the mean squared loss. The two networks' loss functions are defined as follows:

$$L_p = -\log q(\omega_1, \omega_2, ..., \omega_T | I; \pi)$$
$$= -\sum_{t=1}^{T} \log q_\pi(a_t | s_t) \quad (8)$$

$$L_v = ||v_\pi(s_i) - R||^2 \quad (9)$$

Where $L_p$ represents the loss of policy network, $q_\pi(a_t|s_t)$ represents the policy function. When given a certain state $s_t$ at time $t$, the policy function gives a predicted action policy $a_t$.

Secondly, we jointly train the two networks $q_\pi$ and $v_\pi$ using deep reinforcement learning. We learn the parameters of the model by maximizing the total reward that the agent gets when interacting with the environment. We can formulate the reward as $R_t$ which represents the reward at time $t$. The reward expectation is supposed to be: $J(\pi) = \mathbb{E}_{1...T \sim q_\pi}(\sum_{t=1}^{T} R_t)$. For calculation convenience, we can regard the object function of rewards as a Markov decision process. For this problem includes a representation of semantic space from high-dimensional interaction, which may contain unknown environment variables. Considering the solution procedure, we can give an approximation to this problem:

$$\nabla_{q_\pi}^2 J = \sum_{t=1}^{T} \nabla_{q_\pi}^2 \log q_\pi(a_t|s_t)(R - v_\pi(s_t)) \quad (10)$$
$$\nabla_{v_\pi}^2 J = \nabla_{v_\pi}^2 v_\pi(s_t)(R - v_\pi(s_t)) \quad (11)$$
$$v_q(s) = \mathbb{E}[R | a_t \sim q_\pi, s_t = s] \quad (12)$$

Where $v_q(s)$ represents the evaluation scores at state $s_t$ when given a policy predicted by $q_\pi$ at time $t$.

## 4 Experiments

In this section, we perform extensive experiments to evaluate the proposed framework. All the experiments are conducted on the MS COCO dataset, and utilize evaluation indicators such as: BLEU-3, BLEU-4, Meteor, CIDEr and ROUGE-L. All of those evaluation metrics are widely used in caption evaluation tasks.

### 4.1 Dataset Preparation and Network Settings

For the convenience of evaluation, we used the data splits from [4], containing more that 80,000 images for training, 5,000 images for testing and 5,000 images for validation as well.

We use VGG-16 Net [29] as the CNN architecture and LSTM as the RNN architecture. And a three-layer perceptron is used to predict the generated captions to give a reward on the caption. All the inputs are set to be 512 dimensions including the hidden units. And we initialize the model using Adam [37] optimizer with an initial learning rate of $5 \times 10^{-4}$. We adjust the learning rate by a factor of 0.9 every two epochs.

### 4.2 Comparison with the State-of-the-arts

As is shown in Table 1, we provide a summary of our model on MS COCO dataset with five evaluation metrics. Note that the Semantic ATT [7] has utilized rich extra data to train their predictor, so their results are incomparable to the methods without using extra training data. Compared with these methods except [7], our approach shows significant improvements in these evaluation metrics.

**Table 1: Performance of our method on MS COCO dataset compared with state-of-the-art methods. For those competing methods, we show the results from their latest version of paper. The (-) indicates unknown scores.**

| Methods | Evaluation Metric | | | |
|---------|-------|-------|--------|-------|
|         | BLEU3 | BLEU4 | METEOR | CIDEr |
| DeepVS[4] | 32.1 | 23.0 | 19.5 | 66.0 |
| NIC[5] | 32.9 | 27.7 | 23.7 | 85.5 |
| gLSTM[9] | 35.8 | 26.4 | 22.7 | 81.3 |
| m-RNN[11] | 35.0 | 25.0 | - | - |
| ATT[7] | 40.2 | 30.4 | 24.3 | - |
| Ours | 39.5 | 28.2 | 24.3 | 90.7 |

Since our framework has a significant improvement compared to other methods, we'd like to combine the decision-making framework with the existing work to optimize our model constantly in the future work.

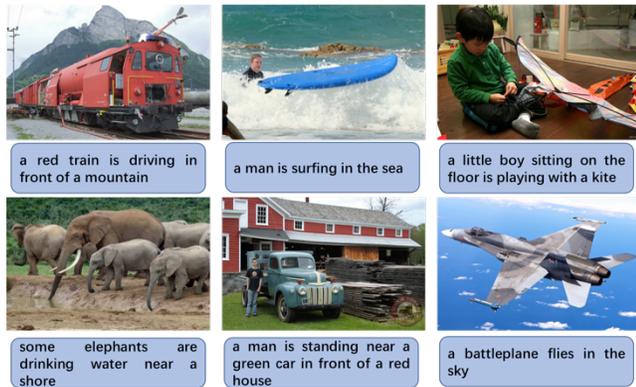

**Figure 4: The visualization results of our model on the MS COCO dataset.**

We show some qualitative captioning results of our model in Fig. 4. From the results, we can see that our method is better at recognizing key objects and the integral environment.

## 5 CONCLUSIONS

In this paper, we present a novel architecture for image captioning with deep reinforcement learning, which can achieve good performance compared with other state-of-the-art methods. Different from the previous encoder-decoder framework, our model utilizes two networks called "policy network" and "value network" to generate captions. The policy network gives some possible actions about the agent. Then, the value network makes decisions on whether to choose the action given by the policy network. And we also use the temporal-difference method to optimize the model. In our future work, we consider to improve the model based on the existing methods and find out other alternative methods to design a new reward mechanism for natural language generation tasks.

## 6 ACKNOWLEDGEMENT

This work is supported by the National Natural Science Foundation of China (No.61501457, No.61602517) and the National Key Research and Development Program of China (Grant 2016YFB0801305).